\documentclass[conference]{IEEEtran}
\IEEEoverridecommandlockouts
\usepackage{amsmath,amssymb,amsfonts}
\usepackage{algorithmic}
\usepackage{graphicx}
\usepackage{textcomp}
\usepackage{xcolor}

\usepackage{tabularx} 
\usepackage{booktabs}
\usepackage[hidelinks]{hyperref}
\usepackage{subcaption}
\usepackage{multirow}
\usepackage{csquotes}
\usepackage{paralist}

\usepackage[backend=biber,style=ieee,sorting=none,sortcites,mincitenames=1,maxcitenames=1]{biblatex}
\usepackage{tikz}
\newcommand\copyrighttext{%
	\footnotesize \textcopyright 2024 IEEE. Personal use of this material is permitted.  Permission from IEEE must be obtained for all other uses, in any current or future media, including reprinting/republishing this material for advertising or promotional purposes, creating new collective works, for resale or redistribution to servers or lists, or reuse of any copyrighted component of this work in other works.}
\newcommand\copyrightnotice{%
	\begin{tikzpicture}[remember picture,overlay]
		\node[anchor=south,yshift=10pt] at (current page.south) {\fbox{\parbox{\dimexpr\textwidth-\fboxsep-\fboxrule\relax}{\copyrighttext}}};
	\end{tikzpicture}%
}
\usepackage{fancyhdr}
\fancypagestyle{firstpage}{%
 \fancyhf{}%
 \fancyhead[C]{Accepted for presentation at 2024 IEEE International Conference on Robotic Computing (IRC)}  
}
\begin{document}

\title{Comparing the Consistency of User Studies Conducted in Simulations and Laboratory Settings
\thanks{This work was partly funded by the German Research Foundation (DFG) under grant agreement HE2696/29 CroViCo.}
}

\author{\IEEEauthorblockN{Jonathan Hümmer}
\IEEEauthorblockA{\textit{Chair for Applied Computer Science III}\\
\textit{(Robotics and Embedded Systems)} \\
\textit{University of Bayreuth}\\
Bayreuth, Germany \\
jonathan.huemmer@uni-bayreuth.de}
\and
\IEEEauthorblockN{Dominik Riedelbauch}
\IEEEauthorblockA{\textit{Chair for Applied Computer Science III} \\
\textit{(Robotics and Embedded Systems)} \\
\textit{University of Bayreuth}\\
Bayreuth, Germany \\
ORCID: 0000-0002-7937-4755
}
\and\IEEEauthorblockN{Dominik Henrich}
\IEEEauthorblockA{\textit{Chair for Applied Computer Science III}\\
\textit{(Robotics and Embedded Systems)} \\
\textit{University of Bayreuth}\\
Bayreuth, Germany \\
dominik.henrich@uni-bayreuth.de}
}
\maketitle
\thispagestyle{firstpage}  
\copyrightnotice
\begin{abstract}
Human-robot collaboration enables highly adaptive co-working. The variety of resulting workflows makes it difficult to measure metrics 
as, e.g. makespans or idle times for multiple systems and tasks in a comparable manner. This issue can be addressed with virtual commissioning, where arbitrary numbers of non-deterministic human-robot workflows in assembly tasks can be simulated. 
To this end, data-driven models of human decisions are needed.
Gathering the required large corpus of data with on-site user studies is quite time-consuming. In comparison, simulation-based studies (e.g., by crowdsourcing) would allow us to access a large pool of study participants with less effort.
To rely on respective study results, human action sequences observed in a browser-based simulation environment must be shown to match those gathered in a laboratory setting. To this end, this work aims to understand to what extent cooperative assembly work in a simulated environment differs from that in an on-site laboratory setting. We show how a simulation environment can be aligned with a laboratory setting in which a robot and a human perform pick-and-place tasks together.
A user study (N=29) indicates that participants' assembly decisions and perception of the situation are consistent across these different environments. 
\end{abstract}

\section{Introduction}
Current research in human-robot collaboration (HRC) for industrial applications emphasizes the need for adaptive approaches rather than adhering to pre-planned, fixed schedules for humans and robots \cite{brunoDynamicTaskClassification2018, riedelbauchDynamicTaskSharing2020}. Such flexibility entails that human and robotic agents make decisions dynamically and coordinate their actions incrementally -- similar to how a purely human team would operate. Such an approach necessitates complex cognitive robotic systems with advanced perception, action, and communication capabilities.

One significant challenge in assessing the benefits of these dynamic interactions is the creation of benchmarks and test protocols that can effectively evaluate largely autonomous processes in a comparable manner \cite{bagchiWorkshopReportTest2020, lenzMechanismsCapabilitiesHuman2014}. Traditional empirical studies, which often focus on quantifying subjective user experiences with systems in controlled laboratory environments, are limited in assessing the dynamic nature of HRC systems: beyond user experience, metrics such as productivity, agent utilization, cycle times, and safety are here crucial for a comprehensive evaluation \cite{riedelbauchBenchmarkingTeamworkHumans2023}. As a consequence of the nondeterminism in human behavior, assessing and comparing these metrics requires analyzing various dynamic processes across multiple scenarios. 
Trying to cover such a variety of workflows with traditional user studies requires extensive temporal, financial, and personnel resources.

Virtual commissioning offers a potential solution to this challenge through simulation-based methods that enable the complete automation of virtual workflow exploration \cite{riedelbauchCognitiveHumanModel2021}. However, a realistic simulation of HRC requires modeling and replicating the decisions and non-determinism of a human interacting with the robot. Again, developing corresponding \emph{cognitive human models} would require more data than feasibly obtained through conventional laboratory-based studies. Therefore, we envision conducting user studies to observe human decisions in assembly tasks online, e.g., by crowdsourcing. From these study results, we will then build human models for virtual cobot system testing. The potential of simulation-based online experimentation has already been demonstrated \cite{anwyl-irvineGorillaOurMidst2020} in the context of HRC \cite{breazealCrowdsourcingHumanrobotInteraction2013, chernovaCrowdsourcingHumanrobotInteraction2011}; however, there is so far no evidence showing that results regarding assembly sequences obtained this way correspond to those one would obtain in an on-site laboratory setting.

\begin{figure*}[t]
    \begin{subfigure}{\columnwidth}
    \includegraphics[width=\linewidth]{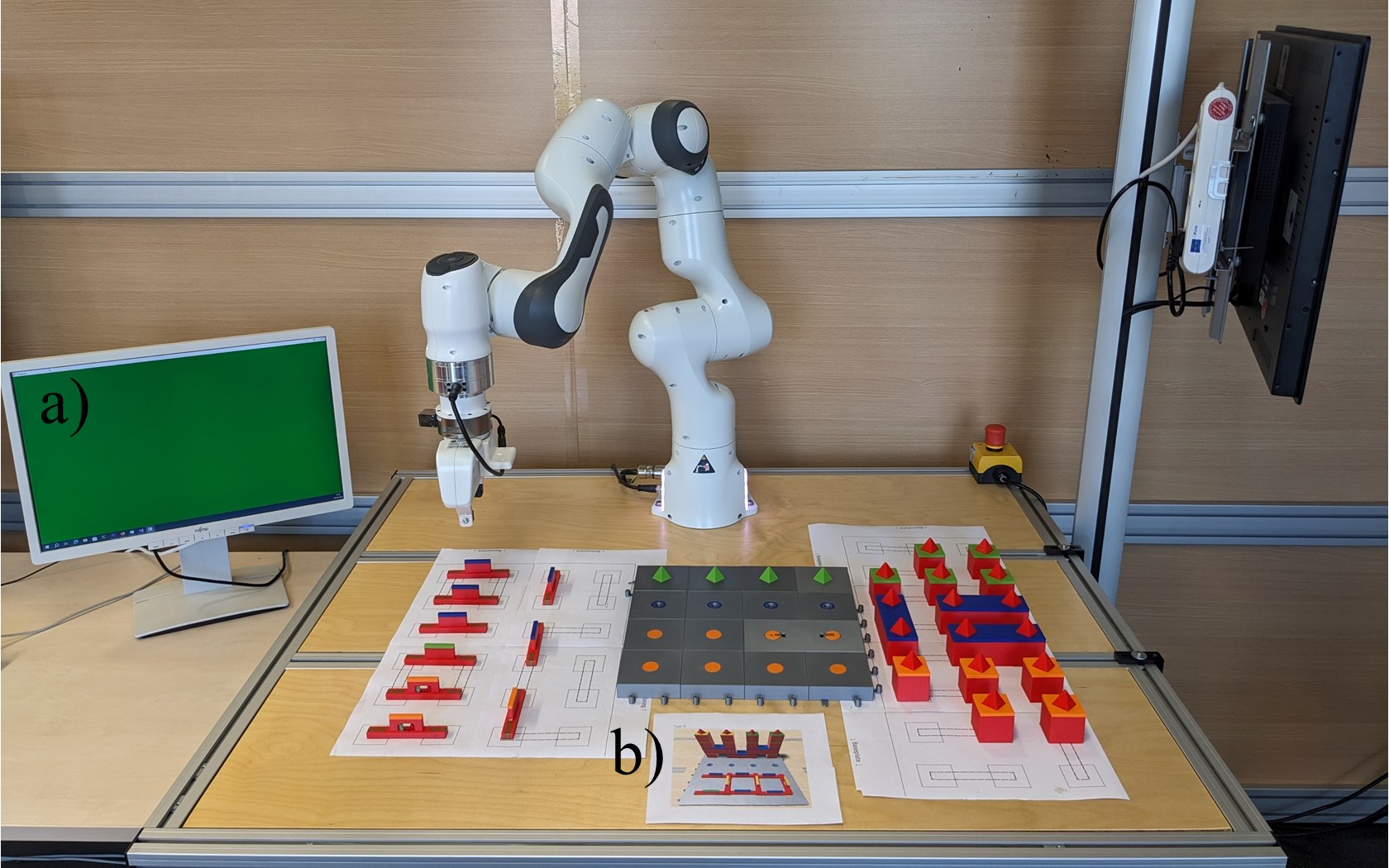}
    \end{subfigure}%
    \hfill
    \begin{subfigure}{\columnwidth}
    \centering
    \includegraphics[width=\linewidth]{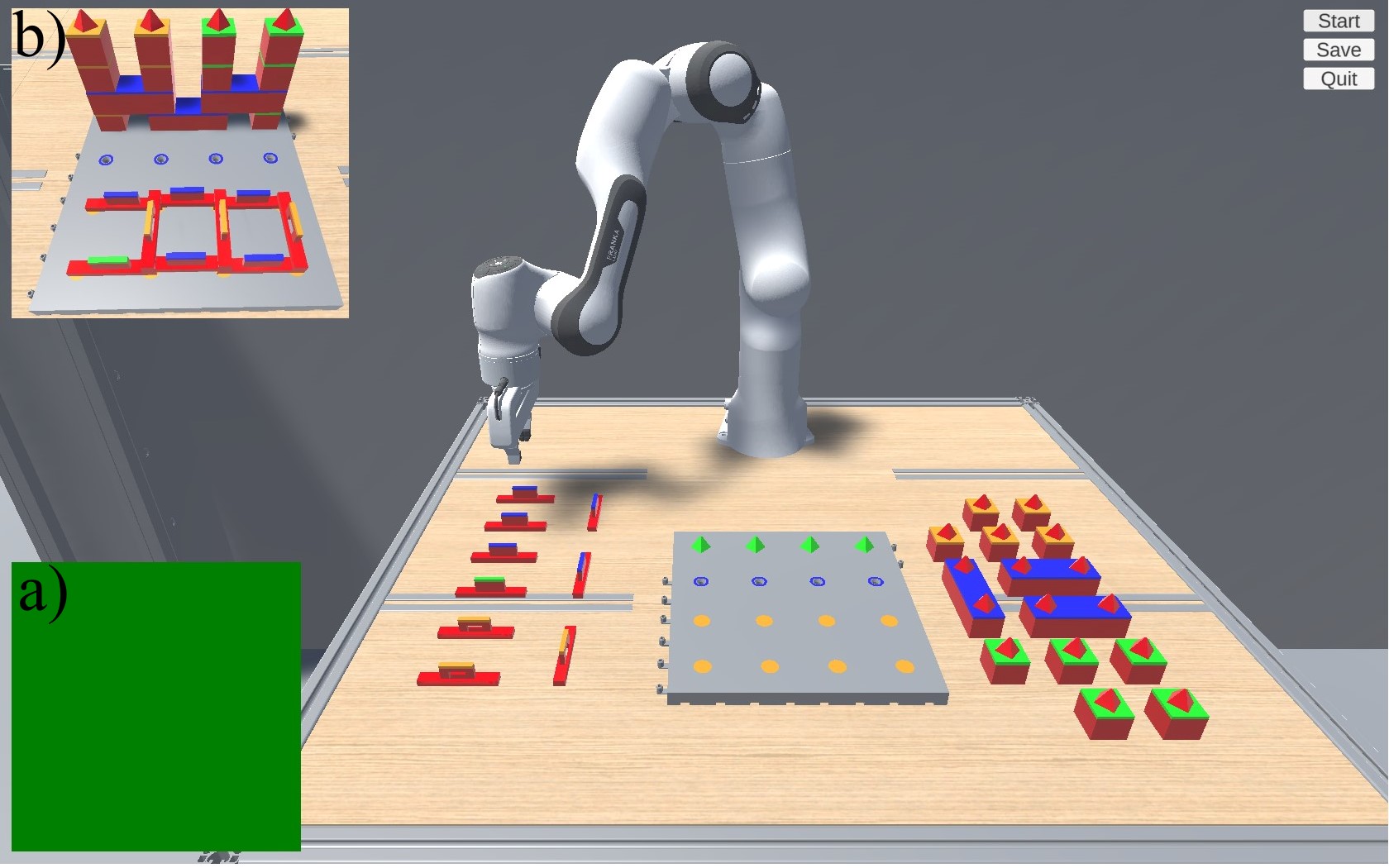}
    \end{subfigure}
    \caption{We compare participants' assembly strategies when solving pick and place tasks with a robot in aligned laboratory (left) and simulation (right) setups. Human speed is regulated by a "traffic light" (a), and identical images show the task (b).\label{fig:Setup lab}}
\end{figure*}

In this paper, we address this gap by conducting a comparative user study in aligned laboratory and simulation environments (\autoref{fig:Setup lab}). We contribute structured considerations on designing an appropriate digital twin (\autoref{sec:methods}). Based on this setup, we secondly show that participants' assembly decisions and perception of the situation are consistent across these different environments (\autoref{sec:results}).

\section{Related Work}
\subsection{User Studies without Physical Robot Setups}
Studies that abstract from traditional laboratory settings are quite established in the wider field of human-robot interaction (HRI) \cite{leeInteractiveVignettesEnabling2021}. Instead of bringing participants in contact with embodied robot systems, corresponding study setups use scenario descriptions ranging from textual descriptions to videos and interactive simulations. Textual descriptions, often presented as so-called vignettes, effectively convey complex scenarios to study participants. They allow for analyzing social behaviors such as trust \cite{lawMeasuringRelationalTrust2020} or communication capabilities \cite{horstmannExpectationsVsActual2020} in HRI without the need for complex laboratory setups. Vignettes have further been employed to explore moral decision-making \cite{komatsuBlamingReluctantRobot2021} and willingness to collaborate with robots \cite{molitorHumanRobotCollaborationSmart2022}. 

Beyond textual representations, images and videos make the examination of dynamic robot behavior and visual robot impact possible. E.g., the influence of human-like robot design on their ethical evaluation has been evaluated with textual vignettes, complemented with photos of robots \cite{willemsEthicsRobotizedPublic2022}. Visual vignettes were further used to investigate preferences for robot anthropomorphism \cite{roeslerWhyContextMatters2022}. Combinations of text and video can also shape participants' expectations for upcoming laboratory studies \cite{eysselEffectsAnticipatedHumanrobot2011}. Specifically, video vignettes have been used to study trust in robots \cite{nessetTransparencyHRITrust2021, kluyWorkingIndustrialCobots2021}, as well as perceived robot emotional intelligence \cite{lawInterplayEmotionalIntelligence2021}.

\subsection{Human-Robot Interaction in Virtual Environments}
The aforementioned studies address how humans perceive robot properties. By contrast, our primary goal is to measure properties of \emph{human-robot teams} (e.g., performance metrics like makespans, fluency, etc. when co-working on a task \cite{riedelbauchBenchmarkingTeamworkHumans2023}). These properties depend on the course of actions when humans and robots work together, and so they require observing the flow of actual human-robot interactions. Simulation environments enable such observations: e.g., \textcite{chernovaCrowdsourcingHumanrobotInteraction2011} have studied human communication behavior by gamification, \textcite{lemaignanSimulationHRIRecent2014} have proposed to test interactive robot navigation with a human in the loop, and \textcite{wollowski17} have used virtual environments to gather data for modeling human behavior. Despite this broad range of possibilities, interactive simulations are still rarely used for user studies, possibly due to the question of consistent results.

\subsection{The question of consistent results}
When conducting studies where users do not experience a robot system directly but through an abstraction like text, images, videos, or a simulation, one must discuss whether the results will be consistent with those obtained when exposing participants to the actual robot. For the case of investigating robot properties relevant to HRI, this question has explicitly been investigated, e.g., by \textcite{willemsEthicsRobotizedPublic2022} and \textcite{Babel21}, who conducted studies both in the laboratory and online with consistent results across both groups. 
To the best of our knowledge, the picture regarding the consistency of joint work on a task in real and virtual environments is less clear. \textcite{chernovaCrowdsourcingHumanrobotInteraction2011} found first indications that human decision-making in simulated environments could align with real-world scenarios. However, the task to solve in their strongly gamified simulation does not match the properties of industrial assembly tasks that we are interested in, i.e., co-working within a small workspace, and sub-tasks with precedence relations -- to close this gap, our contribution is a comparative study to investigate the consistency of human decisions in assembly tasks in laboratory and simulation setups.

\section{Methods}\label{sec:methods}
Our goal is to explore whether simulation environments can replace laboratory-based studies when gathering data on human action decisions. Thus, we aim for participants in a simulation environment to have a perception of the situation that is as similar as possible to the laboratory setting. We took inspiration from the ACT-R model of human cognition \cite{ritterACTRCognitiveArchitecture2019}
to derive potential influencing factors on human behavior and address them during the design and implementation of the laboratory and simulation environment (\autoref{sec:actr}) where possible. However, some differences are inevitable -- we considered these by choosing hypotheses in the study design (\autoref{sec:study_design}) to exclude or confirm them as a source of behavioral differences, hence leading us to co-designed laboratory and simulation setups. 
\begin{figure}[t]
    \centering
    \includegraphics[width=\linewidth]{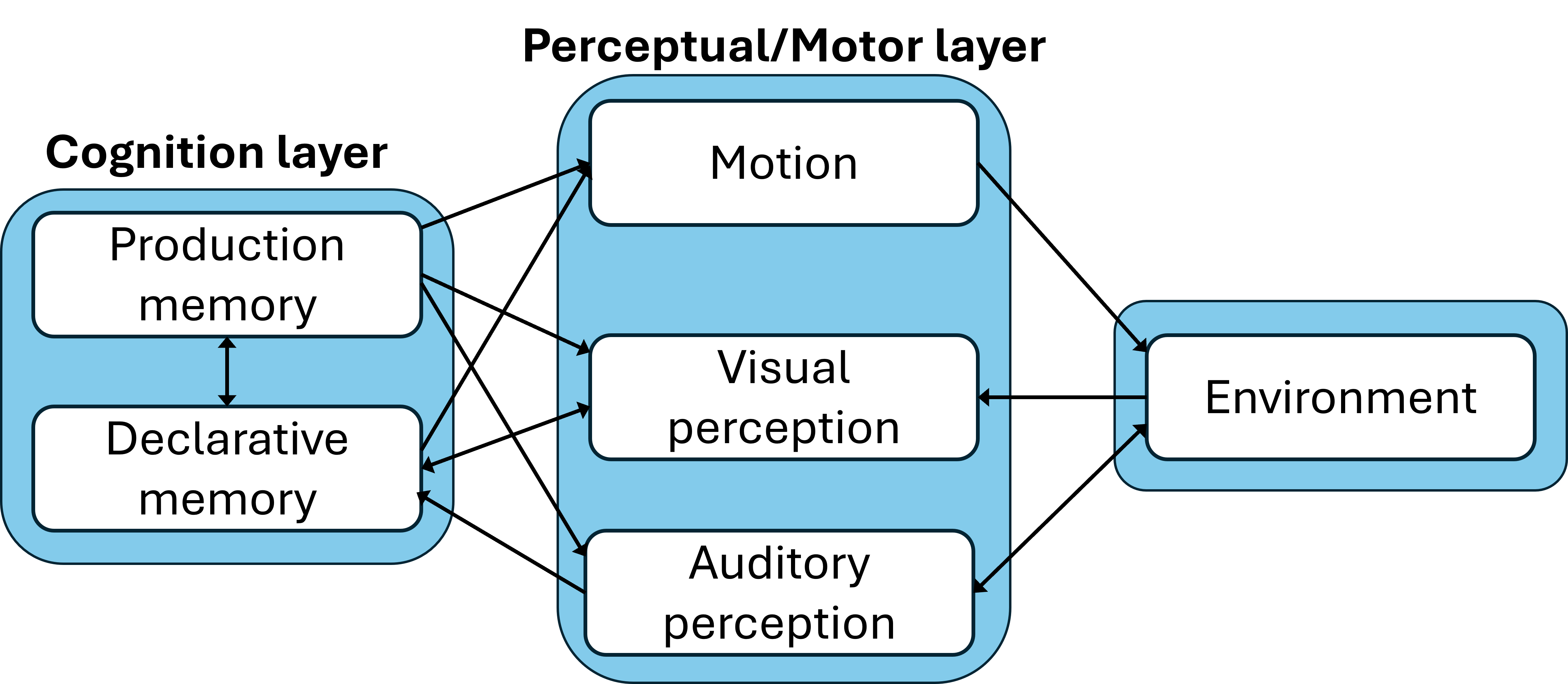}
    \caption{We derive factors influencing human decision-making from the ACT-R model's most commonly used components (adapted from \cite{ritterACTRCognitiveArchitecture2019}).}
    \label{fig:actr}
\end{figure}

\subsection{Alignment of Experimental Setup with the ACT-R Model}\label{sec:actr}
The ACT-R model can be used to predict and explain human behavior \cite{ritterACTRCognitiveArchitecture2019}. In HRI, this model has already been embedded in robot control systems to foresee human team members' behavior \cite{traftonACTREmbodiedCognitive2013}. \autoref{fig:actr} illustrates the main components of the ACT-R model. 
\subsubsection{Production memory} The production memory as a component of the cognition layer consists of "if-then" rules used to select actions based on the state of the other components; thus, in terms of ACT-R, we aim to achieve similar action selection processes in the simulation environment by identifying and minimizing differences to the laboratory setup for all other components to the greatest possible extent.

\subsubsection{Environment}
The environment describes the external world which humans perceive and manipulate. All other ACT-R components either influence or are influenced by the environment. While it may not be feasible to build a simulation indistinguishable from reality, we can still align relevant aspects regarding the task to carry out: We simulate the same robot as used in the lab, and we use the same tasks. Since pick-and-place tasks are frequently used to benchmark HRC \cite{riedelbauchBenchmarkingTeamworkHumans2023}, we focus our investigation on the order in which participants place objects in such a task (\autoref{fig:Setup lab}). As for the objects, we use our synthetic benchmark task environment \cite{riedelbauchBenchmarkToolkitCollaborative2022} -- the objects are designed to compensate for placement inaccuracies of the robot and support reliable study runs without unwanted robot failure. To further increase robustness, we decided to use a graphical user interface where the experimenter inputs the participants' actions instead of relying on a camera for object detection by the robot. This prevents object detection issues related to partial occlusions of the scene, calibration, etc.

\subsubsection{Declarative Memory}
Within ACT-R, the declarative memory is a representation of human factual knowledge, such as information about the current state of an assembly process or the product to be assembled. Differences in behavior could here arise from a different understanding of the goal to be achieved. To prevent this, we show participants identical rendered images of the assembly to complete (e.g., \autoref{fig:tasks}) in simulation and lab setup alike. This task description is continuously available for reference during the process. 
However, we cannot change that humans know by experience that there is no collision risk in the simulation setting. This could lead to humans maintaining a larger distance from the robot in the lab condition, thus possibly influencing action decisions.

\subsubsection{Motion}
The motion component of the perceptual/motor layer models the motor skills of a human.
In our browser-based simulation, participants move parts by "drag-and-drop" with a computer mouse, whereas participants in the lab must physically manipulate the components. In general, it is not possible to avoid this difference in motion patterns directly. However, it indirectly yields a difference in the time human actions take, which we can avoid: during a pilot study, we found that dragging and dropping objects in the simulation was faster than picking and placing them in the the lab setup. Assuming identical speeds of robot motion, this would necessarily mean a confounding factor for action sequences in the simulation since humans would here always take over larger parts of a task than in the laboratory. To establish the same human-to-robot relative working speed in both settings, we introduced a traffic light mechanism. Participants were instructed to only pick a new part when the traffic light shows green, and wait otherwise. We eliminated the difference in human working speeds by adjusting the delay until the traffic light turned green again after placing a part. We moreover used the traffic light mechanism to compensate for the generally longer action durations of cobots when operating at safe speeds. By slowing down human actions, we can provoke more cooperative task completion and better observe the effects of robot actions on human decisions as one part of the study.

\subsubsection{Visual Perception}
Some variations in visual perception of the environment are inevitable due to limitations of rendering in a browser-based environment (cf., e.g., approximations regarding illumination). Moreover, we decided to use a fixed camera perspective in the simulation to simplify simulation control. Despite using a fixed perspective which closely resembles that of a person standing in front of the laboratory workplace, 
this could lead to partial occlusions of the workspace due to robot motion, which can be resolved in the laboratory by moving and changing one's viewpoint. When parts are occluded, this might influence human action decisions. We, therefore, address potential confounds due to this simulation design decision using one of our hypotheses.

\subsubsection{Auditory perception}
Auditory aspects, e.g., natural language interaction of humans and robots, are left to future work. Further environmental factors, such as background noise and the surroundings during the study, arise from the different locations of participants during the study (at a cobot workplace vs. at a PC), and are inevitable. 

\subsection{User Study Design}\label{sec:study_design}

\subsubsection{Hypotheses}
Our first hypothesis H1 directly concerns the primary question of whether action sequences differ between laboratory and simulation. Given that the simulation is designed to replicate the laboratory task and setup closely (Sections \ref{sec:actr}), we hypothesize that participants' assembly decisions will be similar across both settings:
\begin{displayquote}
    \emph{\textbf{H1:} Participants assemble parts with the same strategies in the simulation and the laboratory setup.}
\end{displayquote}

Beyond this main hypothesis H1, we want to explicitly investigate whether participants have a similar working experience under both conditions with further supporting hypotheses. Due to the different modes of perceiving the robot (via a monitor vs. in physical proximity), we first want to question the perceived safety of participants across the two settings:
\begin{displayquote}
    \emph{\textbf{H2:} Participants in the simulation perceive the robot as safe as participants in the laboratory.}
\end{displayquote}
In the simulation, participants sit at a PC and use a mouse to interact with parts, whereas they perform pick and place movements while standing at an assembly station in the laboratory. Due to the lightweight parts and physical ease of the task, we hypothesize that there is still no difference in the physical strain they perceive:
\begin{displayquote}
    \emph{\textbf{H3.1:} Perceived physical workload in the simulation environment is equal to the laboratory setup.}
\end{displayquote} 
The tasks in the simulation are designed to mirror the laboratory scenario precisely, and users receive the same instructions to complete the task in both settings. We, therefore, do not anticipate significant differences in mental workload:
\begin{displayquote}
    \emph{\textbf{H3.2:} Perceived mental workload in the simulation does not differ from the laboratory setup.}
\end{displayquote} 
Lastly, hypothesis H4 is motivated by the static camera perspective in the simulation. Unlike in the laboratory, participants cannot move to change their perspective. Due to the small workspace and alignment of the simulation perspective with typical viewpoints in the laboratory, we assume this limitation to not significantly influence human assembly decisions:
\begin{displayquote}
    \emph{\textbf{H4:} Participants in the simulation are not hindered by additional occlusions caused by the robot as seen from the fixed simulation viewpoint.}
\end{displayquote} 

\subsubsection{Independent variable}
To test these hypotheses, we consider the following independent variable: whether participants work in the laboratory setup or the simulation environment.
We opted for a between-group design to assign participants to the simulation (group \textsc{Sim}) or laboratory setting (group \textsc{Lab}). This design aligns with our vision of replacing traditional laboratory studies with simulation-based online crowdsourcing in the future. The laboratory group is intended to serve as an independent control group to see whether participants in the simulation group act differently, so we want to prevent any confounding factors due to prior experience with the online group or vice versa. 

\begin{table}
	\centering
	\caption{Trust questionnaire for hypothesis H2}
	\label{tab:mapping}
	\footnotesize
    \renewcommand{\arraystretch}{1.1}
	\begin{tabularx}{\linewidth}{@{}p{.2cm}X}
		\toprule
        \multicolumn{2}{@{}l}{\textbf{Questionnaire items (based on \textcite{charalambousDevelopmentScaleEvaluate2016})}} \\
        \midrule
		A & The way the robot moved made me uncomfortable. \\
        B & The speed of the gripper picking up and releasing components made me uneasy. \\
		C & I trusted that the robot was safe to cooperate with. \\
		D & I was comfortable the robot would not hurt me. \\
		E & The size of the robot did not intimidate me. \\
		F & I felt safe interacting with the robot. \\
        \bottomrule
	\end{tabularx}
\end{table}
\begin{table}
	\centering
	\caption{Questionnaire for H3}
    \label{tab:occlusion_questionnaire}
    \footnotesize
    \renewcommand{\arraystretch}{1.1}
	\begin{tabularx}{\linewidth}{@{}p{.2cm}X}
		\toprule
        \multicolumn{2}{@{}l}{\textbf{Questionnaire items}} \\
        \midrule
		G & Did the presence of the robot disrupt or disturb your work experience? \\
        H & Did the robot's actions cause occlusion or block your view of important elements in your work environment? \\
        \bottomrule
	\end{tabularx}	
\end{table}

\subsubsection{Dependent variables}
We measure the following dependent variables to evaluate our hypotheses:

\textbf{H1}: During the user study, we observe the order in which participants transfer parts to goal positions which we internally identify with a unique letter each (\autoref{fig:tasks}). An assembly sequence is then encoded by a string. Individual assembly sequence strings can have different lengths because participants work with a robot that also performs operations in the task in an adaptive way. Given these strings, we use the Levenshtein distance \cite{levenshtein1966binary} to compare the similarity of assembly sequences. This measure counts the minimum number of insertions, deletions, or substitutions to transform two strings into one another. This allows us to cluster action sequences even if they have different lengths.

\textbf{H2:} Regarding perceived safety, we rely on the trust questionnaire designed for industrial human-robot collaboration proposed by \textcite{charalambousDevelopmentScaleEvaluate2016}, since it has already been applied to simulation-based studies previously \cite{kluyWorkingIndustrialCobots2021}. Given our focus on perceived safety in this study, we narrowed the original ten questionnaire items down to the items listed in \autoref{tab:mapping} -- the other items primarily address the perceived reliability of robot components (particularly the gripper), and so they are less relevant for the purpose of H1. All items are scored on a five-point Likert scale. 

\textbf{H3:} We use the NASA Task Load Index (NASA TLX) \cite{hartNasaTaskLoadIndex2006} for the hypotheses directed towards perceived workload. Here, we are particularly interested in the raw subscale scores for physical (H3.1) and mental workload (H3.2). We kept the remaining subscales in the questionnaire for exploratory purposes, and we asked participants to score the subscales on 11-point Likert scales.

\textbf{H4:} For H3 concerning occlusions caused by the robot, we designed a custom questionnaire due to a lack of standardized measures for this special case. In this questionnaire,  workflow disruptions due to occlusions are assessed through two questions to be answered on five-point Likert scales (\autoref{tab:occlusion_questionnaire}).

\section{Results}\label{sec:results}

\subsection{Study Procedure}\label{sec:procedure}
The comparative study was conducted in three phases: 
\subsubsection{Briefing}
Participants first read written instructions on cooperative interaction with the system (particularly the semantics of the stoplight), and descriptions of the tasks to complete. 
In the laboratory study, participants could briefly familiarize themselves with the robot by touching and moving it in compliant control mode. To ensure that participants understood the instructions, they were required to complete a short tutorial assembly without the robot.

\subsubsection{Task Execution}
\begin{figure}[t]
	\begin{subfigure}[t]{0.413\linewidth}
		\includegraphics[width=\linewidth]{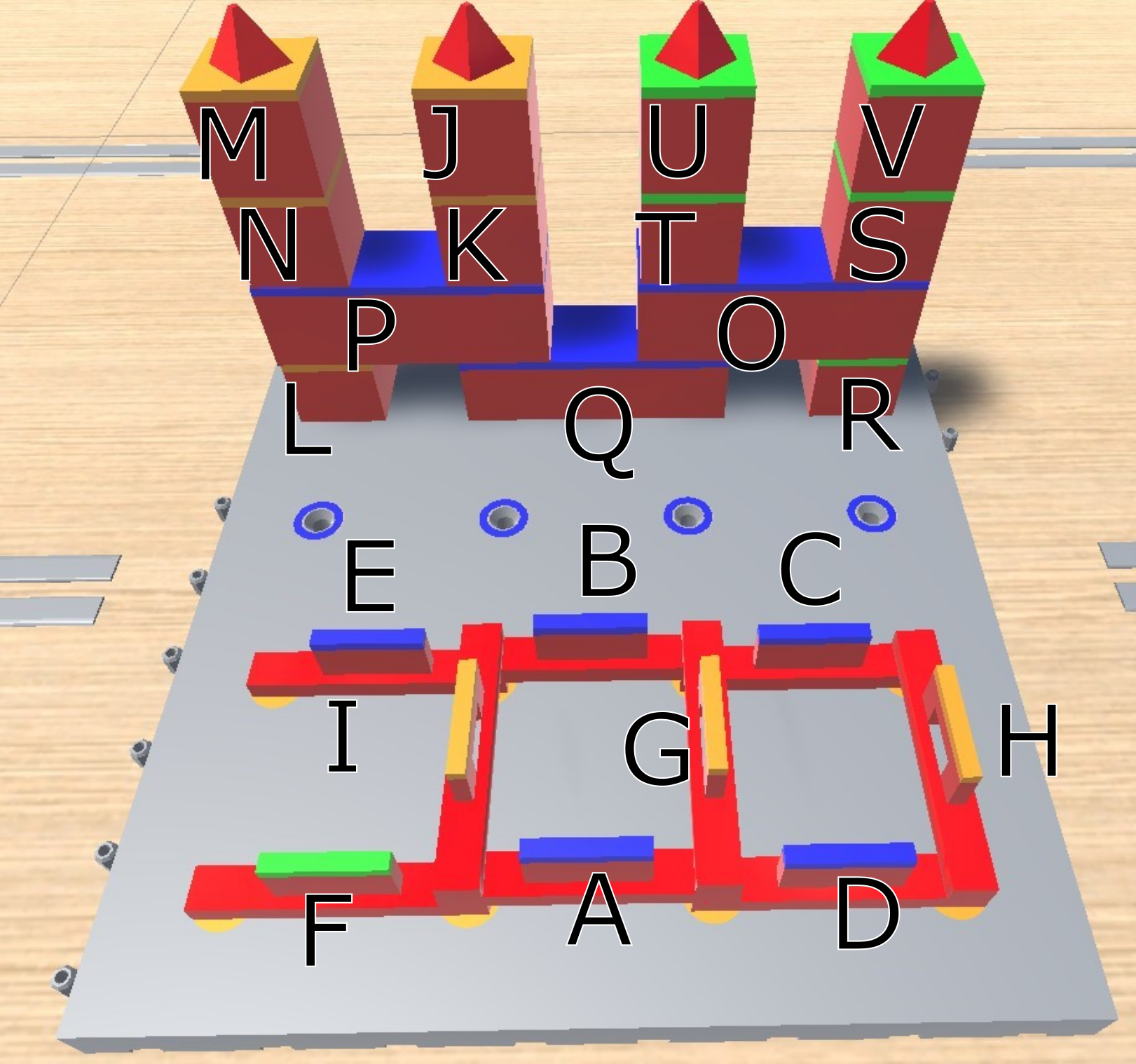}
		\caption{Task 1}
		\label{fig:task1}
	\end{subfigure}
	\begin{subfigure}[t]{0.567\linewidth}
		\hfill
		\includegraphics[width=\linewidth]{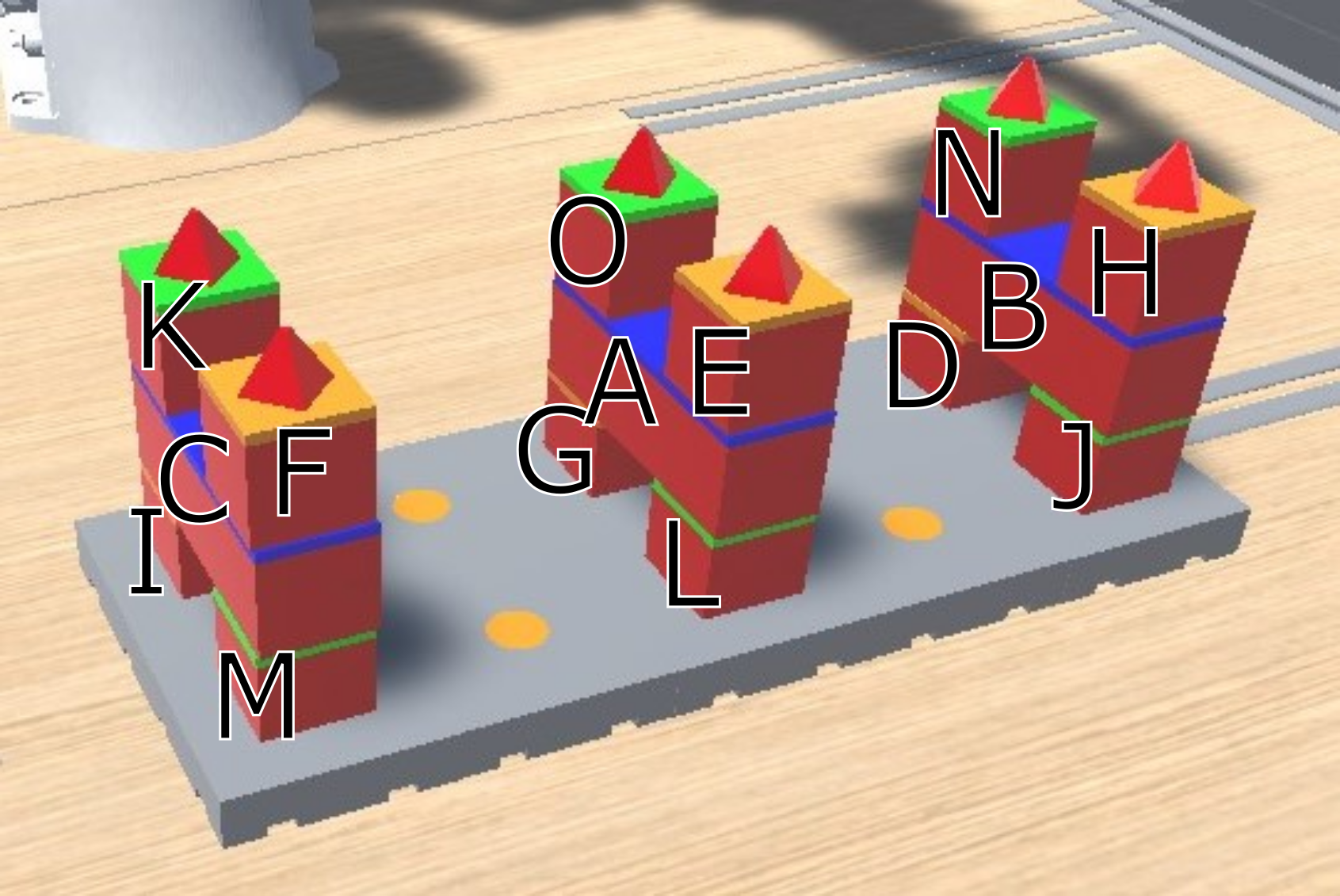}
		\caption{Task 2}
		\label{fig:task2}
	\end{subfigure}
	\caption{The descriptions of the main study tasks. To derive assembly sequences each action received a label. During the study, however, participants were shown unlabeled images.\vspace{-2.5mm}}\label{fig:tasks}
\end{figure}

Participants then proceeded to the two main tasks (\autoref{fig:tasks}). Task 1 involves assembling two substructures, whereas Task 2 requires building three identical towers. Both tasks allow participants considerable freedom in placing components, as multiple correct positions are available for each piece. This opens the working process to different action sequences, also due to interference with actions performed by the robot. 
After examining the task descriptions, participants were given the opportunity to ask questions. For the \textsc{Lab} condition, participants worked at the workstation shown in \autoref{fig:Setup lab} (left). Work in the simulation environment was performed either remotely or in a seminar room if we had to provide hardware.

\subsubsection{Post-study questionnaires} After completing the tasks, participants answered a demographic questionnaire and the questionnaires as outlined in \autoref{sec:study_design}. 

\subsection{Participants}
The study was conducted in the robotics laboratory at the University of Bayreuth with 29 participants. We accepted participant registrations consecutively, and each new participant was assigned to the group \textsc{Sim} or \textsc{Lab} at random. Participants' ages ranged from 19 to 52 years, with a mean (M) of 25.7 years (standard deviation SD = 5.9). Of these participants, 16 (55.2\%) identified as male, whereas 13 (44.8\%) identified as female. A post-study analysis showed that age and gender were evenly distributed across the two groups.
Participants in the \textsc{Sim} group received a compensation of 5€, while the \textsc{Lab} group received 10€ due to longer setup times in the laboratory.

\subsection{Statistical Evaluation}
Studies on HRC are commonly focused on proving differences between conditions. E.g., researchers would usually want to show that a specific scheduling algorithm $A$ resulted in less workload for human participants than some other approach $B$. This is commonly done by statistical hypothesis testing: with typical approaches as the $t$-test, we can formally show that the null hypothesis $H^0$ dual to the investigated hypothesis $H$ ($H^0$: \textquote{Algorithms $A$ and $B$ yield the same workload}) must be rejected if the $p$-value is below a pre-defined confidence threshold. Then, rejecting $H^0$ means accepting the original hypothesis on this confidence level.

In our case, however, we want to investigate whether the two conditions \textsc{Lab} and \textsc{Sim} result in the same human behavior, i.e., instead of proofing an effect, we are seeking to proof \emph{absence} of an effect. Regarding H1, the Levenshtein distance provides us with an objective measure to prove the similarity of assembly sequences. However, to show the absence of an effect for our hypotheses H2 to H4 with classical hypothesis testing, we would need to conduct an equivalence trial to reject the dual null hypothesis that behavior is different. In theory, this can be achieved with the "two one-sided tests" (TOST) strategy. However, TOST requires defining the smallest effect size of interest \cite{harrisNotStatisticallyDifferent2012}. As opposed to, e.g., clinical trials, we are currently not aware of a solid way to achieve this for our setting and dependent measures.

Since we can thus not directly test whether working conditions are perceived identically across conditions, we will only use H2 to H4 to support H1 as follows: We treat H2 to H4 as null hypotheses, i.e., we will try to prove a statistically significant difference instead of trying to prove the absence of a difference for perceived safety, workload, etc. If we find that we have to reject H1, and if we find a significant difference regarding any of the other hypotheses, then this difference will provide us with a possible reason for the difference in assembly sequences; this could inform future work to improve the similarity of the \textsc{Lab} and \textsc{Sim} setup. Vice versa, if we find similar assembly sequences and accept H1 while still finding a difference in any of H2 to H4, then we will take this as an indication of a Type II error regarding H1. 
This methodology has a specific limitation: with fewer samples, it becomes more likely not to detect significant differences even though they exist. However, we consider our sample size of 29 to be sufficient for identifying such differences, as this sample size is typical for laboratory studies in HRC \cite{zimmerman22}.

We applied the Shapiro-Wilk test and found that participants' questionnaire answers were mostly not normally distributed ($p < 0.5$, \autoref{tab:trust stats}). To seek statistically significant differences between the two groups as outlined above, we thus used the non-parametric Mann-Whitney-U test (MWU).
\begin{table}[t!]
	\caption{Statistical test results}\label{tab:trust stats}
	\footnotesize
	\centering
	\begin{tabularx}{\linewidth}{@{}lXXl@{}}
    \toprule
	& \multicolumn{2}{l}{\textbf{Shapiro-Wilk}} & \multirow{2}{*}{\textbf{MWU}} \\
	& \textsc{Lab} & \textsc{Sim} &  \\
    \midrule
    \multicolumn{4}{@{}l}{\textbf{Trust questionnaire} (see \autoref{tab:mapping})} \\
    \midrule
	\text{A} & \phantom{\textless}0.004 & \textless 0.001 & 0.070 \\
	\text{B} & \textless 0.001 & \textless 0.001 & 0.731 \\
	\text{C} & \textless 0.001 & \textless 0.001 & 0.810 \\
	\text{D} & \textless 0.001 & \textless 0.001 & 0.705 \\
	\text{E} & \textless 0.001 & \phantom{\textless}0.001 & 0.146 \\
	\text{F} & \textless 0.001 & \textless 0.001 & 0.112\\
    \midrule
    \multicolumn{4}{@{}l}{\textbf{Questions regarding occlusions} (see \autoref{tab:occlusion_questionnaire})} \\
    \midrule
        \text{G} & \phantom{\textless}0.018 & \phantom{\textless}0.001 & 0.250\\
        \text{H} & \phantom{\textless}0.045 &\phantom{\textless}\textbf{0.140} & 0.812\\
    \midrule
    \multicolumn{4}{@{}l}{\textbf{NASA Task Load Index} \cite{hartNasaTaskLoadIndex2006}}\\
    \midrule
        Mental Demand & \phantom{\textless}0.007 &  \textless 0.001 & 0.224 \\
		Physical Demand & \phantom{\textless}0.007 &  \textless 0.001 & 0.241 \\
		Temporal Demand & \phantom{\textless}0.002 & \phantom{\textless}0.042 & 1.000 \\
		Performance & \phantom{\textless}\textbf{0.094} & \phantom{\textless}0.006 & 0.928 \\
		Effort &  \textless 0.001 & \phantom{\textless}0.006 & 0.164 \\
		Frustration &  \textless 0.001 &  \textless 0.001 & 0.489\\
    \bottomrule
    \end{tabularx}
\end{table}
\begin{figure*}[h!]
	\begin{subfigure}[t]{\columnwidth}
		\includegraphics[width=\linewidth]{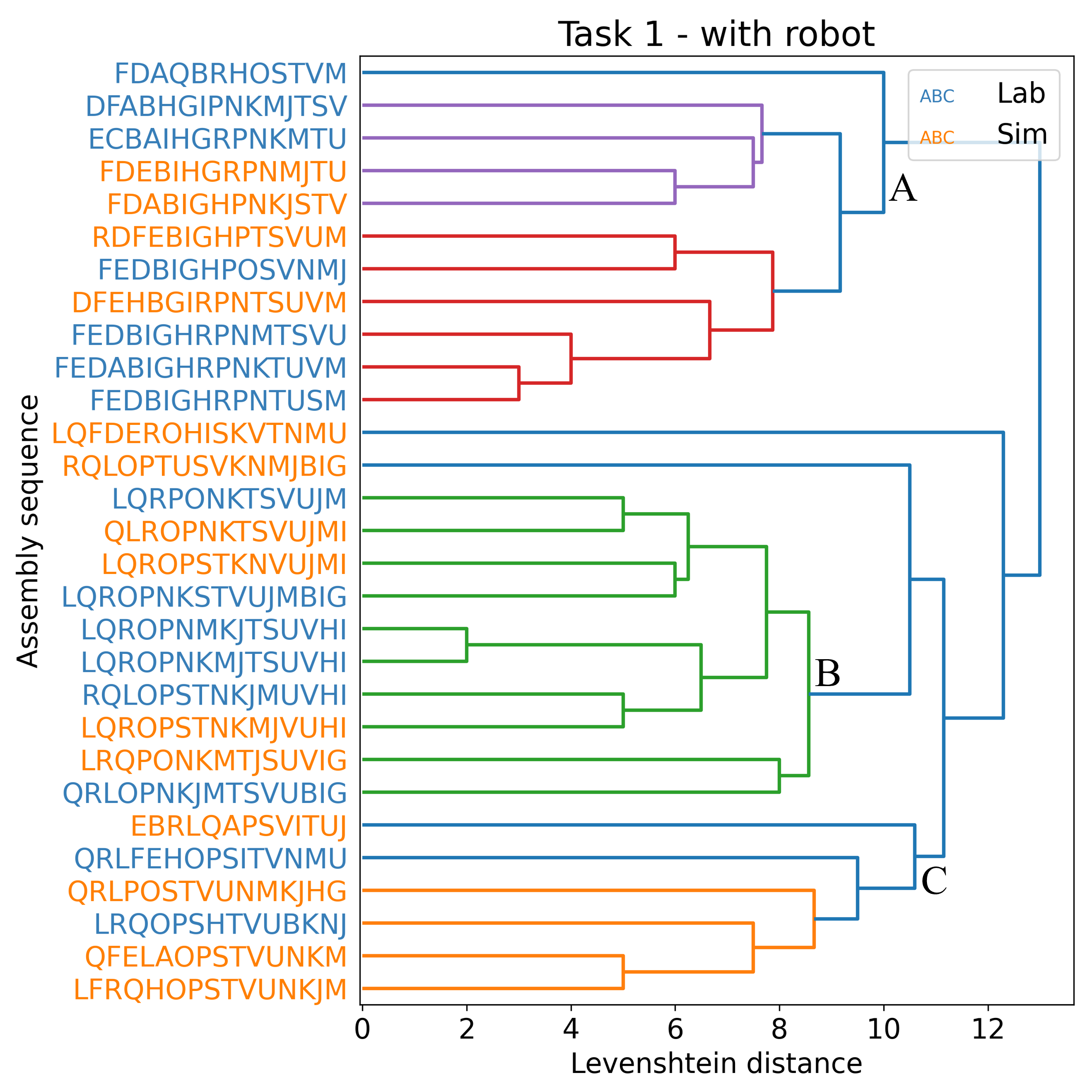}
	\end{subfigure}%
     \hfill
	\begin{subfigure}[t]{\columnwidth}
		\includegraphics[width=\linewidth]{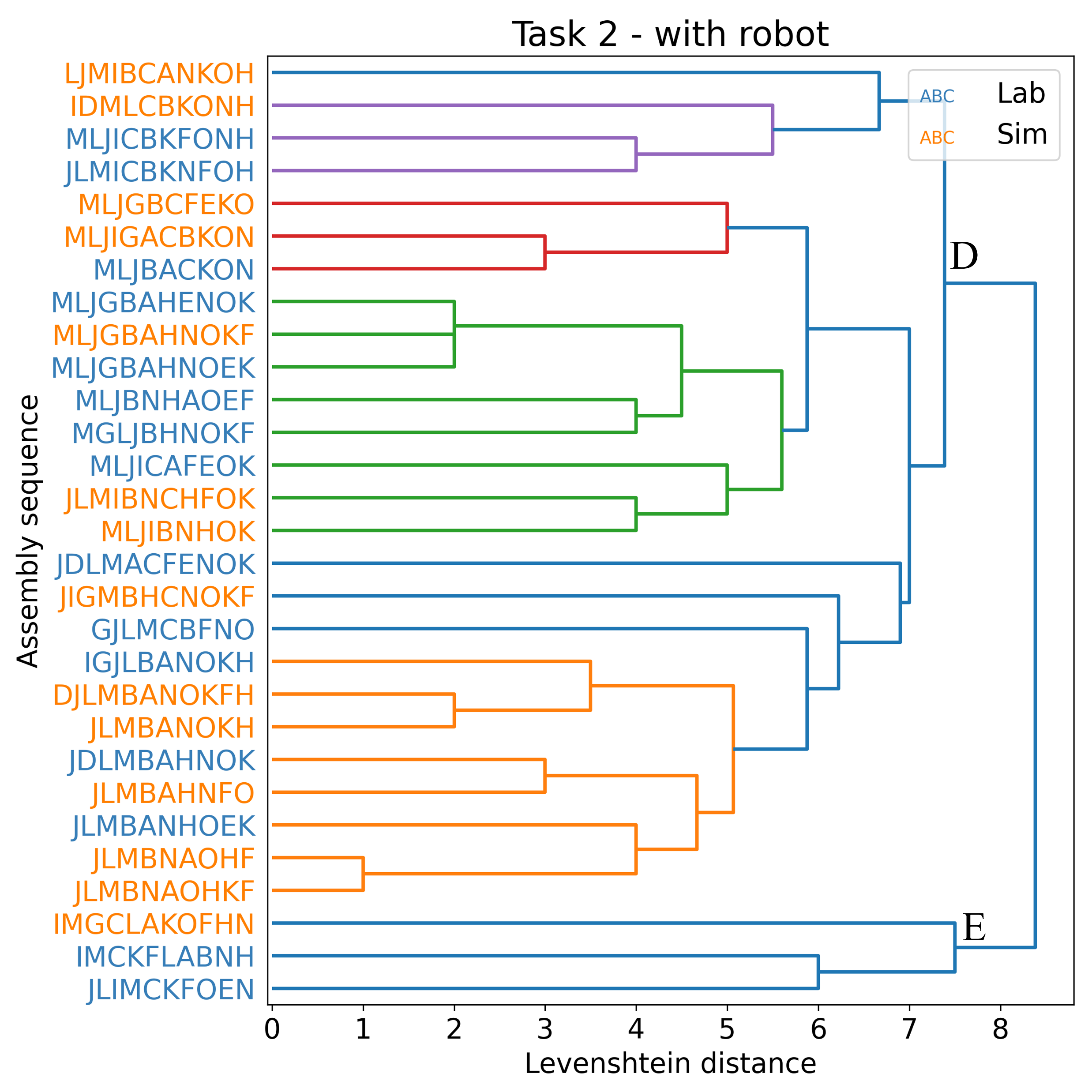}
	\end{subfigure}
	\caption{The observed assembly sequences can be grouped into clusters (A, B, C, D, E) indicating aligned working strategies. Individual strategies were observed for participants in the \textsc{Lab} and the \textsc{Sim} condition alike.}\label{fig:sequences}
\end{figure*}

\subsubsection{H1 (consistency of assembly sequences)}
We clustered the observed assembly sequences hierarchically according to their Levenshtein distance. For each cluster, we could identify a consistent and plausible assembly strategy that is matched by all sequences in that cluster: in Task 1 (\autoref{fig:task1}), eleven participants (four from the \textsc{Sim} group) began by constructing the front substructure first (cluster A, \autoref{fig:sequences}). Another group of ten participants (six from the \textsc{Sim} group) started with the rear substructure using cubes (cluster B). Eight participants (six from the \textsc{Sim} group) alternated between adding parts to the two substructures (cluster C). 
For Task 2 (\autoref{fig:task2}), the majority of 26 participants (13 from the \textsc{Sim} group) built the structure layer by layer (cluster D). Only three participants (two from the \textsc{Sim} group) preferred to finish one of the tower-like sub-structures before proceeding to the next. 
Interestingly, this trend towards layer-by-layer work has previously also been observed by \textcite{mayerCognitiveEngineeringAutomated2011}.

\subsubsection{H2 (perceived safety)}
On average, participants felt quite safe with the robot in both settings (\autoref{fig:safety}). Across all questionnaire items, we did not find any statistically significant differences in answer distributions with the MWU test ($p>0.05$, \autoref{tab:trust stats}). This is surprising for us since we have identified potential fear of collisions as an inevitable difference between \textsc{Sim} and \text{lab}\footnote{Indeed, the MWU test result for Item A ("The way the robot moved made me uncomfortable.") is close to the significance threshold.} -- we hypothesize that the familiarization of participants in the \textsc{Lab} (\autoref{sec:procedure}) has contributed to this result. 
It could also be a result of the convenience sampling: 40\% of participants in the \textsc{Lab} group already had prior experience working with robots, which could have reduced their fear of collisions. 
One person in the \textsc{Lab} group felt very uncomfortable with the way the robot moved (Item A). Moreover, one person in the \textsc{Lab} and one in the \textsc{Sim} group were intimidated by the size of the robot (Item E). Due to the otherwise consistent results, we consider these answers to be outliers.

\subsubsection{H3 (perceived workload)}
Participants assigned the tasks an overall low perceived workload (\autoref{fig:tlx}). The MWU test results do not indicate a statistically significant ($p>0.05$) difference for any of the TLX dimensions (\autoref{tab:trust stats}). In line with our expectation due to the measures taken (\autoref{sec:study_design}), and with the consistent assembly strategies we observed, we conclude that we successfully built a simulation environment that maintains all key aspects of task work in the laboratory.

\subsubsection{H4 (occlusions)}
\begin{figure}
    \begin{subfigure}[t]{\columnwidth}
        \includegraphics[width=\linewidth]{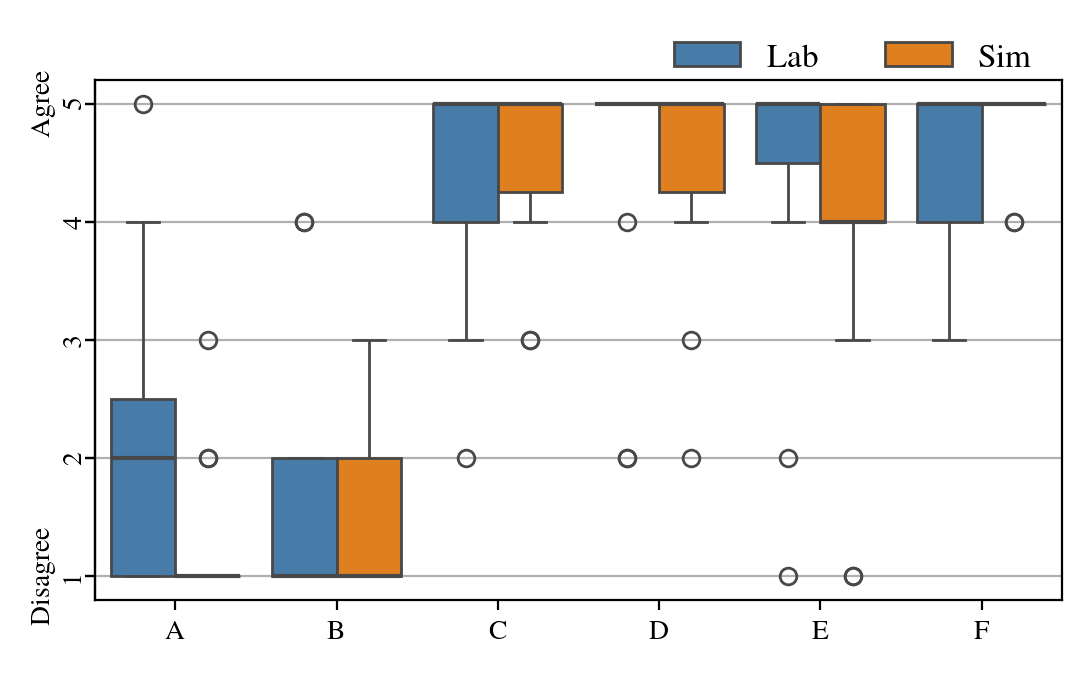}
		\caption{Perceived safety (see \autoref{tab:mapping})}\label{fig:safety}
    \end{subfigure}
    \begin{subfigure}[t]{\columnwidth}
        \includegraphics[width=\linewidth]{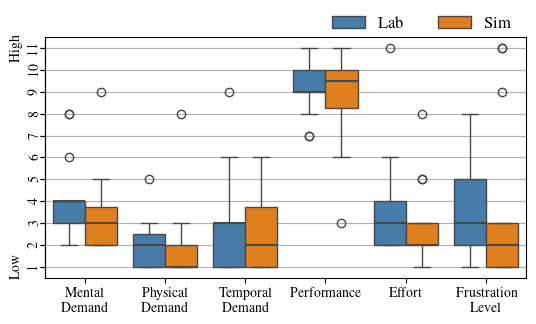}
		\caption{NASA TLX}\label{fig:tlx}
    \end{subfigure}
    \begin{subfigure}[t]{\columnwidth}
        \includegraphics[width=\linewidth]{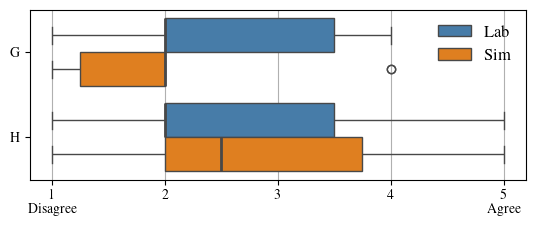}
        \caption{Occlusion questionnaire (see \autoref{tab:occlusion_questionnaire})}\label{fig:occlusion}
    \end{subfigure}
    \caption{Post-study questionnaire results}
\end{figure}

In the \textsc{Lab} group (M=2.5, SD=1.1), most participants did not feel disrupted by the robot in their work experience according to our occlusion questionnaire (\autoref{fig:occlusion}). On average, participants in the \textsc{Sim} group felt even less disrupted (M=2.0, SD=1.0). The amount of perceived occlusions was rated neutrally and with high variance in both the \textsc{Lab} (M=2.6, SD=1.2) and the \textsc{Sim} group (M=2.7, SD=1.4). We did not find statistically significant differences ($p>0.05$) between the two groups with the MWU test. These results indicate that participants in both the \textsc{Lab} and \textsc{Sim} environments had similar perceptions of disturbances and occlusions due to the robot. This also suggests that the static camera perspective in the simulation did not significantly influence their decisions. However, the high variance of the answers indicates much noise in the data. 
One potential source is our custom questionnaire, which requires further investigation. 

\subsection{Conclusion}
All in all, we draw the following conclusion from the above results: our goal was to show that human assembly strategies for identical tasks are consistent, no matter whether the task was conducted in a physical laboratory environment or in the virtual simulation setting (H1). To this end, we observed assembly sequences in both settings and found that they can be grouped into consistent strategies followed by participants in the \textsc{Lab} and \textsc{Sim} settings alike. All strategies were followed by participants in both groups, i.e., we did not find a strategy unique to either \textsc{Lab} or \textsc{Sim}.
Furthermore, each cluster consists of mostly equal proportions of participants from both groups.
However, for some strategies in Task 1, the number of participants was slightly imbalanced between the two settings (particularly in clusters A and C). We attribute this to statistical effects and the sample size rather than taking it as an indication to reject H1 for the following reasons: based on our analysis in \autoref{sec:actr}, possible causes for differing behavior due to differently perceived safety, workload, or disruptions due to robot actions, are covered by the supporting hypotheses H2 to H4. The statistical evaluation has not shown any statistically significant differences between \textsc{Lab} and \textsc{Sim} regarding these factors. We acknowledge that this absence of proof for differences does not necessarily prove that participants perceived the situation identically \cite{harrisNotStatisticallyDifferent2012} -- still, we argue that a difference in behavior must have a cause in underlying cognitive processes. Hence, it is unlikely that a difference in behavior would come without a measurable difference in the underlying cognitive aspects we tested. 
In consequence, we consider the big picture drawn by these joint observations regarding H1 and H2 to H4 as strong evidence that our main hypothesis H1 is valid.

\section{Summary and Outlook}
Virtual commissioning of HRC systems requires the development of data-driven cognitive human models to replicate nondeterministic human action decisions. In this paper, we have presented evidence that the data required for such models can be gathered by observing human behavior likewise in a browser-based simulation environment or with on-site lab studies. To this end, we used the ACT-R model of human cognition to co-design a laboratory as well as a simulation study while avoiding factors that might yield differences in action selection. A between-group comparative study (N=29) has shown no statistically significant differences between work in the laboratory vs. the simulated environment regarding perceived safety, workload, or obstructions due to limitations of simulation implementation. Moreover, the assembly decisions of participants have been shown to exhibit similar patterns -- hence, we conclude that simulated environments are a valid method for observing human decisions in assembly tasks.

In our future work, we are planning to validate the approach with larger sample sizes from online crowdsourcing. This will enable us to gather data for a variety of tasks. We also intend to address the issue of equivalence testing of perceived trust and workload between \textsc{Lab} and \textsc{Sim} setting to directly validate consistency of results.
Moreover, we will investigate suitable machine-learning models to embed the observed behavior patterns into virtual human models for HRC benchmarking.

\printbibliography
\end{document}